# A data-driven approach to discover and quantify systemic lupus erythematosus etiological heterogeneity from electronic health records

Marco Barbero Mota[1,†], John M. Still[1], Jorge L. Gamboa[2], Eric V. Strobl[3], Charles M. Stein[2], Vivian K. Kawai[2], Thomas A. Lasko[1,4,*]

Vanderbilt University Medical Center, Department of [1]Biomedical Informatics, [2]Medicine, [3]Psychiatry and Behavioral Sciences and [4]Computer Science

[†]Vanderbilt DBMI PhD program, 2525 West End Ave, Nashville, TN; *Primary advisor

**Abstract.** *Systemic lupus erythematosus (SLE) is a complex heterogeneous disease with many manifestational facets. We propose a data-driven approach to discover probabilistic independent sources from multimodal imperfect EHR data. These sources represent exogenous variables in the data generation process causal graph that estimate latent root causes of the presence of SLE in the health record. We objectively evaluated the sources against the original variables from which they were discovered by training supervised models to discriminate SLE from negative health records using a reduced set of labelled instances. We found 19 predictive sources with high clinical validity and whose EHR signatures define independent factors of SLE heterogeneity. Using the sources as input patient data representation enables models to provide with rich explanations that better capture the clinical reasons why a particular record is (not) an SLE case. Providers may be willing to trade patient-level interpretability for discrimination especially in challenging cases.*

## Introduction

Systemic lupus erythematosus (SLE) is a complex relapsing disease that manifests through various combinations of symptoms and clinical signs. SLE's heterogeneity makes its recognition in the health record slow and subjective. SLE diagnosis is usually reached after excluding other likely explanations to patient symptoms[1].

Current practice relies on the application of imperfect classification criteria by non-specialists to trigger a subsequent referral to a rheumatologist. These tools have traditionally focused on either sensitivity or specificity at the expense of the other[2] with the most recent revision balancing both metrics[3]. These criteria were designed to form uniform phenotypic cohorts for research, but they disregard the high variability of clinical presentations among SLE patients[4]. Some ineligible patients are at high risk of being missed or misdiagnosed, a phenomena referred as spectrum effect[5] that leads to high false negative rates[1,4,6]. These include patients at the early stages of the condition where some clinical signs may have not yet developed, single organ/system-dominant forms, negative antinuclear antibody (ANA) test cases, and rare but severe presentations[7].

SLE's heterogeneity is also a reason why few new drugs have been successfully developed. The current one-size-fits-all approach applied in other autoimmune diseases such as rheumatoid arthritis (RA) has not been efficacious for SLE[8,9]. Even in positive trials, only marginal treatment effects have been found which highlights the need to dissect patient populations into clinically homogeneous subgroups which would increase clinical trial efficiency[9].

Delimiting patient subgroups at scale requires pulling apart the components that make up SLE's heterogeneity. Clinically meaningful sources of variation must link manifestational criteria and the underlying causal mechanisms of disease (i.e. etiology)[10]. Identifying the independent causes of SLE diagnosis may allow to accurately define targets for novel therapies in more efficient trials[11,12].

Some SLE subtypes have been recognized in the literature as constellations of non-specific clinical manifestations and laboratory values[1,6]. A recent study proposed a model that divided SLE into type 1 and 2[13,14]. There also exist established SLE subtypes due to their significant differences in prognosis and treatment such as lupus nephritis[15] or lupus with antiphospholipid syndrome[16]. However, these have been defined based on clinical experience and have neither been quantitively characterized nor causally validated.

Electronic health records (EHR) are a rich source of medical information routinely collected in clinical practice. Prior research has leveraged EHR data to develop phenotyping algorithms that attempt to recognize SLE records at scale: expert-guided decision trees[17] and supervised machine learning (ML) models trained on chart review[18,19] or noisy[20] labels. A recent study also incorporated genetic data and found no gain over EHR data alone[21]. All these systems were optimized for accuracy, which places greater emphasis on unsurprising easy cases at the expense of the long tail of less common SLE facets, a phenomenon known as hidden stratification[22]. The patterns these algorithms uncover are likely not causal which precludes models to fully capture the data generating process. Using causal predictors enhances model interpretability which may be preferred to better accuracy in recognizing heterogenous diseases[23,24].

In this study, we demonstrate that large amounts of health records data and independence-based pattern discovery[25] is sufficient to identify the imprints left in the EHR by the independent *latent sources* of the SLE label which describe the disease's heterogeneity. The disentanglement of these patterns or *signatures* is done in a completely unsupervised manner without any prior knowledge: we *let the data tell us* what the signatures are. Under a set of causal assumptions, the corresponding independent sources have been shown to represent the root causes of the data generating process[26]. Lasko et al. showed that using this causal representation of patient data as input to ML models was more accurate than the original clinical variables in predicting the malignancy of solitary lung nodules. Notably, the most predictive latent sources appeared to represent undiagnosed cancer in some patients[25]. In this work we do not try to forecast future disease but rather recognize SLE in the EHR across its clinical facets. We also probe whether the signatures of predictive sources can provide inference into SLE etiological heterogeneity.

In this work we make the following **contributions**:
i) We infer 2000 probabilistically independent latent sources and their EHR signatures from noisy, sparse, and irregular data from a broad sample of rheumatology patients.
ii) We evaluate the benefits of using the sources to recognize SLE in the health record, when compared to using estimates of the observed clinical variables at the same points in time.
iii) We identify 19 probabilistically independent sources predictive of SLE whose EHR signatures describe clinically recognizable pictures of SLE heterogeneity.
iv) We demonstrate the higher expressivity and interpretability that the sources bring to supervised models as they allow to quantify the patient-level root causes of SLE being present in the health record.

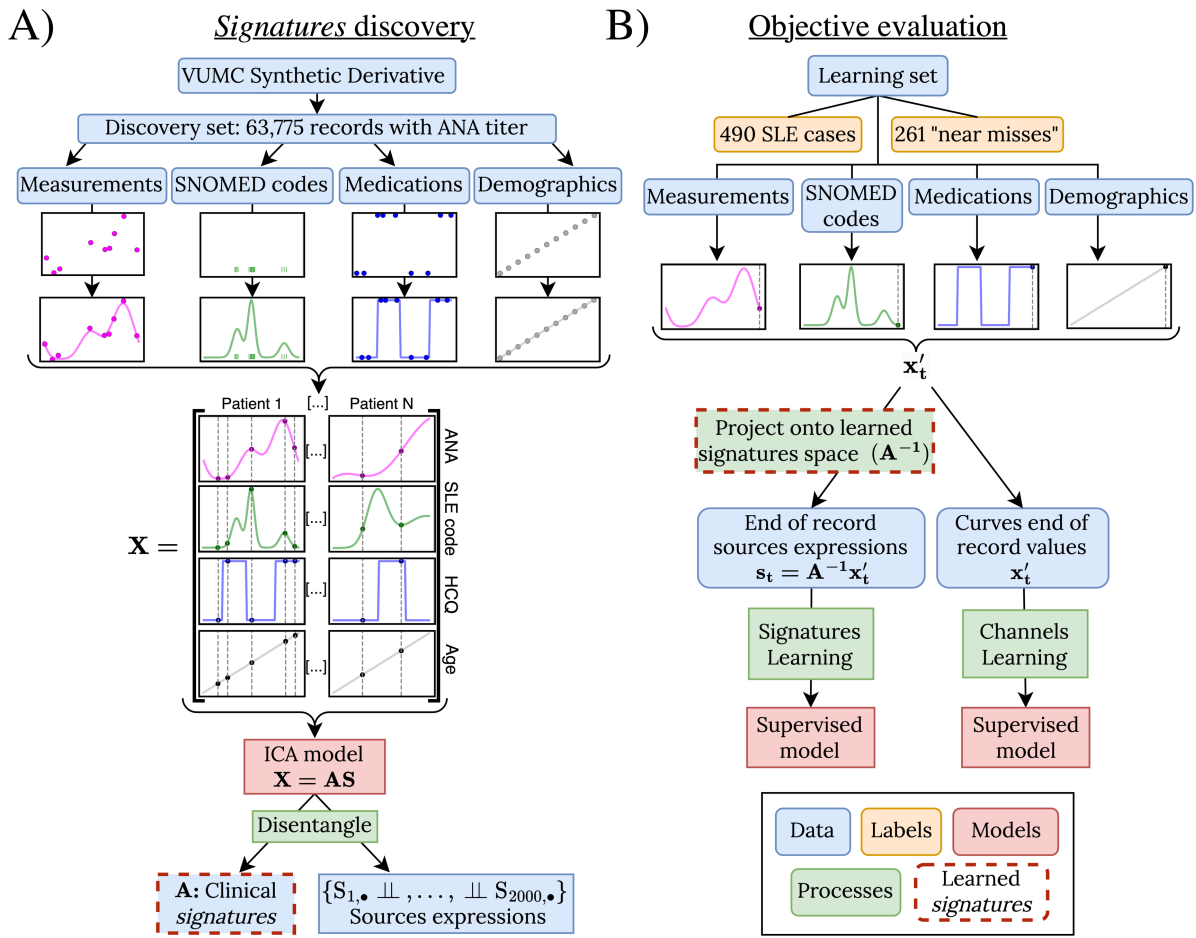

**Figure 1.** Data flow summary. **A)** Noisy, sparse observations are used to generate longitudinal curves, one per clinical variable (channel) and per patient. For each individual, all channels curves are sampled at random times. Cross-sections for the discovery set are aggregated into a dense data matrix ($\mathbf{X}$) that is decomposed into 2000 approximately independent sources expressions at the input timepoints and their EHR signatures. **B)** Supervised ML models are trained to discriminate SLE cases from 'near misses' (negative) records. Learning happens from either the channels curves last value or their projections onto the signatures space (i.e. the sources expressions levels).

**Methods**
This study was conducted at Vanderbilt University Medical Center (VUMC) and was determined by the VUMC's IRB to be non-human-subjects research.

**Data.** We extracted all data from VUMC's de-identified EHR mirror, the Synthetic Derivative[27] (SD), which hosts longitudinal information for over 3 million patients. With cutoff date 02/16/2018, we collected historical patient data for patients who had had an ANA test done, with or without a titer result. This cohort conforms our discovery set and considers a wide spectrum of autoimmune and non-autoimmune conditions.

We collected data from diagnosis codes, medications, clinical measurements and demographic information (race, age, and biological sex) variables which we refer to as channels. We discarded those with less than 1000 total events or appearing in less than 10 records in the discovery set. We mapped SNOMED concepts from ICD-9/10 codes.

**Data denoising, synchronization and fusion.** We followed Lasko et al.[25] for EHR data preprocessing and clinical signature discovery (figure 1A) and refer to the paper for full details. In summary, we computed longitudinal curves for all considered channels with daily resolution from the discrete asynchronized noisy observations. Each data modality follows a specific curve generation process:

a) Conditions (SNOMED codes): We inferred smooth code-intensity curves using a variation of Random Average Shifted Histograms[28] that accounts for the non-stationarity of the event-arrival distribution. In the event of missing data, we imputed a constant curve with a baseline annual event-arrival probability of 1/20.
b) Clinical measurements (laboratory values and BMI): We used the univariate monotonic cubic interpolator PCHIP[29] to generate smooth curves without overshooting. Records without observations for a measurement were imputed a constant curve with the population median.
c) Medications (RxNorm): Medication data only exists when noted in a visit's medication list and missing otherwise. We built binary curves through a best-guess approximation of drug-taking between visits. We considered that a regime was continued between contiguous visits if noted on both. On the contrary, if a particular medication was missing at the later visit, the midpoint between dates was considered as stopping time. We imputed a zero constant curve for those records that did not include any mention for a particular drug.
d) Demographics: Race and sex were constant binary variables. Age curves were linearly increasing integers.

Unlike Lasko et al.[25], we did not perform post-computation curve smoothing. Instead, we directly stacked each patient curves longitudinally to form a $p \times \tau$ curveset, where $p$ is the number of clinical variables and $\tau$ is the number of days spanned by the record. Next, we serially concatenated all curvesets and performed $n$ cross-sections at uniformly random times with an average density of one sample per record-year. Merging of all data produced a dense data matrix $\mathbf{X} \in \mathbb{R}^{p \times n}$ whose columns represent synchronized complete instances of patient records. We standardized $\mathbf{X}$ variable-wise to bring all variables into roughly the same scale. Measurements, log-transformed condition codes intensities, and age were mean centered and scaled by two standard deviations[30]. Binary variables (medications, race and biological sex) were left untouched[30].

**Signature discovery.** We use probabilistic independence as our guiding principle to discover clinical signatures from $\mathbf{X}$ (figure 1A). This choice comes naturally from the assumption that latent mechanisms of disease (or sources) operate independently and leave a pattern imprint (or signature) in the clinical history of a patient[25,26,31]. We formalized our unsupervised decomposition through an Independent Component Analysis (ICA) model[32]:

$$\mathbf{X} = \mathbf{AS} . \qquad (1)$$

For implementation we used the fastICA algorithm[33] (Python 3.10, scikit-learn 1.1.3) to obtain $\mathbf{A} \in \mathbb{R}^{p \times k}$ and $\mathbf{S} \in \mathbb{R}^{k \times n}$ which correspond to the mixing and source matrices, respectively (figure 1). Each column of $\mathbf{A}$ ($\mathbf{A}_{\bullet,i}$) depicts a signature, or the $p$ linear mixing weights that describe the trace left in the record by the corresponding latent source ($\mathbf{S}_{i,\bullet}$). The $j$-th column of $\mathbf{S}$ ($\mathbf{S}_{\bullet,j}$) quantifies the sources expressions for the $j$-th data instance ($\mathbf{X}_{\bullet,j}$), or how active each source was in the $j$-th cross-section. We set the number of latent sources $k = 2000$ due to RAM memory constraints.

We made a first approach to evaluating the signatures clinical face validity by subjectively assessing their resemblance to common clinical pictures. For such task we used the signatures description diagrams (see Results figure 5) that quantitatively display the signature ($\mathbf{A}_{\bullet,i}$) and the histogram of its expressions ($\mathbf{S}_{i,\bullet}$) across the discovery set. All signatures are given a random sequential identifier by the algorithm (e.g. S-650). We provide with descriptive names to those that are relevant for the discussion of recognizing SLE in the health record.

**Objective signatures evaluation.** We evaluated the disentangled sources by quantifying their predictive power in recognizing SLE health records among a labelled set of patients or learning set (figure 1B). The learning set included the records in our discovery set with at least one SLE ICD-9/10 code, no codes for systemic sclerosis and/or dermatomyositis, and with existing EHR-linked genomic data who were exclusively of European ancestry (i.e. white race). Patients from the learning set were assigned a binary label through chart review by three SLE experts (JG, VK and CMS) who followed the definition of SLE case used by Barnado et al.[17]: An SLE case must contain an explicit diagnosis by a VUMC or external rheumatologist, dermatologist or nephrologist in the clinical notes. Patients with only cutaneous lupus and drug-induced lupus were not considered SLE cases. This definition included both easy and hard cases to diagnose. Negative records were labelled as "near misses" due to their historical suspicion of SLE.

We extracted the last available cross-section of the channels curves ($\mathbf{x}'_t$) for all learning set patients and projected them onto the signatures space following the learned ICA model (equation 1):

$$\mathbf{s}_t = \mathbf{A}^{-1} \mathbf{x}'_t . \tag{2}$$

$\mathbf{x}'_t$ and $\mathbf{s}_t$ are therefore two representations of the same information in the channels and signatures space respectively. We used these datasets as input to train four ML architectures to discriminate SLE cases from "near misses". We trained Elastic Net (ENet)[34] (scikit-learn 0.24.2, python 3.9.12) and Adaptive Elastic Net (AdaNet)[35] equipped with sure independence screening[36] (in-house implementation on python 3.9.12) penalized logistic regression. These architectures introduce L2 and L1 penalties in their loss functions which provides with stability against collinearity and automatic feature selection, respectively. ENet penalties exhibit the grouping effect which leaves either in or out of the model sets of strongly correlated predictors. This behavior may zero out important features and include noise variables into the model. Under a set of weak regularity conditions, the AdaNet overcomes this problem as it achieves the oracle property guaranteeing variable selection consistency[35].

Despite their simplicity, linear models may be less interpretable since they may not fully capture the true data generating function[24]. Decision trees ensembles are flexible non-linear models that tend to outperform linear architectures in many supervised tasks. We trained Random Forest[37] (scikit-learn 0.24.2, python 3.9.12) and XGBoost[38] (xgboost 1.7.4, python 3.9.12) to account for any non-linearity that may persist the ICA transformation. We made a random 70/30 split of our learning set to form training and test sets. We tuned each model architecture under a two-step process. First, we randomly searched across the hyperparameter space using Bayesian optimization[39] (optuna 3.1.0, python 3.9.12) with 10-fold cross-validation area under the receiver operating characteristic curve (10f-CV AUROC) as target and a universal computational budget of 4000 trials. Since the variance of K-fold CV may be very large[40], we employed a second step to increase tuning robustness. From the initial search results, we identified the subset of hyperparameters combinations that were statistically comparable in performance with the best model through overlap of DeLong[41,42] logistic[43] confidence intervals (CI) ($\alpha = 0.2$). We then optimized for mean out-of-bag AUROC on 100 bootstrapped samples to find the final model training AUROC for which we report Wald 95% CIs. This tuning process was performed for all architectures in both the signatures and the channels. To select the final model architecture, we computed test set performance of all four architectures on both data representations and quantified statistical significance of their pairwise differences via the DeLong test ($\alpha = 0.05$) for correlated AUROCs.

For each combination of model architecture and data representation, we also computed the integrated calibration index (ICI)[44] and cross-entropy loss along their pivot-based 95% CIs based on 1000 bootstrapped samples. The former quantifies the accuracy of individual predictions while the latter combines both calibration and discrimination.

We benchmarked our two final models against the SLE phenotyping algorithm with the highest PPV (95%) in Barnado et al[17] by computing its recall, specificity, and precision on our test set.

**Feature importance.** We trained our final signature model under 500 bootstrapped samples of the training set to obtain the empirical distribution of global feature importance, computed as the mean absolute SHapley Additive exPlanations (SHAP)[45] values across the test set. We used linear SHAP[45] and assumed feature independence to stay true to the model[46] and align with the causal definition of SHAP[47]. The independence assumption is automatically satisfied over the signatures model predictors under the ICA model but does not hold true for the channels.

**Record-level root causes.** Figure 2A describes the causal data generation process of this study problem of recognizing SLE in the health record. The disease process, its causes (e.g. genetics and environmental exposures), and its effects (e.g. pathophysiologic or lifestyle changes), are the most upstream latent nodes. These cause a set of disease observations ($O_D$) that are recorded by a clinician. The provider may then recommend the patient a certain treatment that she may or may not adhere to ($T$). We do not observe the actual treatment but only what the clinician documents in the EHR ($O_T$). The actual treatment causes a set of downstream effects to which we also do not have direct access ($DE$) (e.g. drug secondary effects). $T$, $O_T$, and $DE$ go on to cause additional observations in the record by the clinician ($O_{DE}$). As described above, a binary chart review label is assigned to each patient by examining the record (i.e. $O_D$, $O_T$ and $O_{DE}$, which include the clinical note that reflects the physician diagnosis).

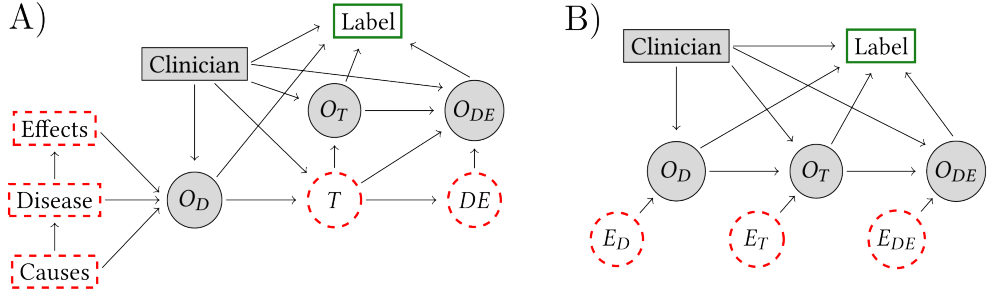

**Figure 2.** Two views of the causal graph that describe the data generating process for a particular patient. **A)** Includes conceptual latent variables while in **B)** unobserved nodes are compressed into error terms. These are root causes to the degree observed data allows for upstream identification. Shaded grey nodes correspond to observed dimensions, dashed lines define latent nodes, and the green box represent the binary SLE label. D: Disease; O: Observations; T: Treatment; DE: Downstream effects; E: Error term.

The predictive true root causes behind the individual patient label are the parentless latent nodes in figure 2A. In an ideal world, we would like to observe and use these as input to predictive models. However, we only have access to imperfect observations of them[48]. Under the LiNGAM model assumptions[49], latent sources discovered using probabilistic independence represent unobserved root causes of the target sink node in the data generating process to the extent these causes are identifiable from the observed data[26]. Formally, ICA sources correspond to the exogeneous independent error terms of the structural equation model that we assume to describe our problem causal process (figure 2B). Causal relations are transitive, where A causes C if A causes B and B causes C. The discovery of the error terms from observational data allows for a compact representation of unobserved causes (dashed nodes) in figure 2A using the data we access to (gray rounded nodes). We can thus equivalently represent the data generative process using the causal graph in figure 2B, where we have removed $T$ and $DE$ from Figure 2A without introducing confounding[50].

The sources in matrix $\mathbf{S}$ are equivalent to the predictive error terms ($\mathbf{E}$), each associated to an observed variable[26]. The individual predictivity of each source for each patient mathematically corresponds to its SHAP value ($\phi$) given a model that estimates $P(\text{Label} \mid \mathbf{S})$[26]. If $\phi_i > 0$ then the *i-th* source is a root cause of the SLE label, while if $\phi_i < 0$ the source is protective. Thus, SHAP values provide a quantitative causal explanation of why a patient record did or did not receive an SLE label. We obtain SHAP explanations for the same patient using both the signatures and channels final models to demonstrate the added expressivity and information richness of the sources.

**Results**

**Cohorts and dimensionality.** 63,775 patients fulfilled our inclusion criteria and comprised the discovery set. The final EHR data dimensionality was $p = 7947$, composed of 6218 SNOMED condition codes, 839 clinical measurements, 879 medications, race, age, and biological sex. Curve sampling generated $n = 646{,}775$ cross-sections stacked in $\mathbf{X} \in \mathbb{R}^{7947 \times 646{,}775}$ and fed to ICA for its decomposition into $\mathbf{S} \in \mathbb{R}^{2000 \times 646{,}775}$ and $\mathbf{A} \in \mathbb{R}^{7947 \times 2000}$. The learning set consisted of 490 SLE cases and 261 negative but difficult to diagnose "near misses". Demographics for patients in all these cohorts are shown in table 1.

**Model performance.** There were no significant differences on the training accuracy among architectures for either data representation (table 2). The channels achieved higher training AUROC point estimates than the signatures. On the test set, the channel models were statistically more accurate. No architecture showed statistically better performance which suggests that the underlying function was sufficiently captured by the linear models. Test accuracy was generally higher than in training which aligns with the idea that the test set had a larger share of easy cases to classify. AdaNet and XGBoost showed the best calibration among models trained on the signatures and the channels, respectively. In general, the signatures models were better calibrated, with XGBoost as the exception. Cross-entropy was slightly better for both linear models in using the signatures and for ENet and XGBoost on the channels data representation (figure 3).

We chose the AdaNet as architecture for the rest of the analysis due to its performance and calibration per above as well as its feature selection consistency (oracle property) and higher interpretability (low complexity with fewer nonzero features).

**Table 1.** Demographics for the different patient cohorts. Std: standard deviation.

| Demographic | | Discovery set | Training set | | Test set | |
|---|---|---|---|---|---|---|
| | | | Near misses | SLE cases | Near misses | SLE cases |
| Unique patients count (N) | | 63,775 | 182 | 343 | 79 | 147 |
| Age (years) | Mean | 53 | 57 | 52 | 57 | 50 |
| | Std. | 18 | 15 | 16 | 16 | 15 |
| Sex [%] | Male | 21,086 [33.06] | 29 [15.93] | 39 [11.37] | 7 [8.89] | 17 [11.56] |
| | Female | 42,684 [66.93] | 153 [84.07] | 304 [88.63] | 72 [91.14] | 130 [88.44] |
| | Unknown | 5 [0.008] | 0 [0] | 0 [0] | 0 [0] | 0 [0] |
| Race [%] | White | 51,497 [80.75] | 181 [99.45] | 340 [99.13] | 77 [97.47] | 147 [100] |
| | Black | 7,539 [11.82] | 0 [0] | 0 [0] | 1 [1.27] | 0 [0] |
| | Asian | 957 [1.50] | 0 [0] | 0 [0] | 0 [0] | 0 [0] |
| | Native American | 373 [0.58] | 0 [0] | 1 [0.29] | 0 [0] | 0 [0] |
| | Pacific Islander | 133 [0.21] | 0 [0] | 0 [0] | 0 [0] | 0 [0] |
| | Multiple | 420 [0.66] | 1 [0.55] | 2 [0.58] | 1 [1.27] | 0 [0] |
| | Declined | 49 [0.08] | 0 [0] | 0 [0] | 0 [0] | 0 [0] |
| | Unknown | 2,807 [4.4.9] | 0 [0] | 0 [0] | 0 [0] | 0 [0] |

Our final models outperformed Barnado et al. algorithm[17] on the test set (figure 4A and B) even though the estimation of their performance is probably optimistic. Their work and ours share the same data source but a completely independent train/test split which may have leaked patients from their training set into our test set.

**Predictive sources of SLE's heterogeneity.** The channels model learnt from the full training set selected 8 clinical variables which included SNOMED codes (SLE, SLE with organ/system involvement, chest pain, ECG normal, and low back pain) and laboratory tests (ANA titer, DNA double stranded antibody test and Complement C4).

The final signatures model selected 5 latent sources (figure 4C). S-650 describes a recognizable description of SLE with the particularity that the systemic autoimmunity has advanced to damage a particular organ or system (figure 5), which is a strong predictor of definite SLE diagnosis[6]. S-1289 shows the characteristic picture of a generic SLE patient, slightly anemic female with increased intensity of the unspecific SLE code. S-1497 represents the SLE complication of lupus nephritis[15]. S-1588 depicts typical SLE laboratory abnormalities of high ANA titer, low complements, and elevated SLE-specific DNA double-stranded antibody test[1]. Finally, S-1683 shows a picture of toxic maculopathy in patients with codes for autoimmune diseases. Manual review revealed that this source is highly expressed in records documenting long-term use of hydroxychloroquine among patients with confirmed autoimmune conditions such as SLE, RA and Sjögren's syndrome. S-1683's signature accounts for increments in the probability of exposition to the drug of up to 0.18, however this change is not shown in the diagram as it fell below the display cutoff (figure 5). Manual review of selected clinical notes revealed that providers use the toxic maculopathy code for insurance purposes when recommending hydroxychloroquine users for their annual eye exam. This may explain why the extremely rare disease toxic maculopathy was found to be the top contributor of S-1683. Graphically, S-1683 is an example of $E_{DE}$ in the causal graph (figure 2B) while the other 4 predictive sources are error terms of the set depicted by $E_D$.

**Table 2.** Supervised models performance and calibration. AUROC: Area under the receiver operating characteristic curve (higher is better). ICI: Integrated calibration index (less is better). CE: Cross entropy (less is better). Brackets enclose 95% CIs. Bolding denotes statistically equivalence within columns by pairwise DeLong test ($\alpha = 0.05$). *: Random Forest and XGBoost AUROC difference was significant.

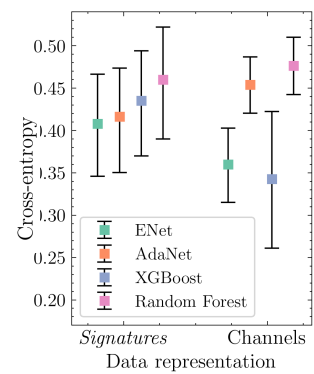

**Figure 3.** Test cross-entropy.

| Model | Training AUROC | | Test AUROC | | Test ICI | | Test CE | |
|---|---|---|---|---|---|---|---|---|
| | Signatures | Channels | Signatures | Channels | Signatures | Channels | Signatures | Channels |
| ENet | 0.88 [0.83, 0.91] | 0.91 [0.87, 0.94] | **0.89 [0.84, 0.93]** | **0.93 [0.89, 0.96]** | 6.4 [2.8, 8.8]×10⁻² | 8.9 [5.4, 12]×10⁻² | 0.41 [0.35, 0.47] | 0.36 [0.32, 0.40] |
| AdaNet | 0.87 [0.82, 0.90] | 0.90 [0.86, 0.94] | **0.88 [0.83, 0.92]** | **0.92 [0.88, 0.95]** | 4.9 [0.8, 6.5]×10⁻² | 12 [8.5, 14]×10⁻² | 0.42 [0.35, 0.48] | 0.45 [0.42, 0.49] |
| XGBoost | 0.85 [0.79, 0.89] | 0.91 [0.87, 0.94] | **0.88 [0.82, 0.92]** | **0.92 [0.88, 0.95]** | 6.8 [3.1, 9.6]×10⁻² | 3.5 [0.0, 4.2]×10⁻² | 0.44 [0.37, 0.49] | 0.34 [0.26, 0.42] |
| Random Forest | 0.84 [0.77, 0.88] | 0.91 [0.87, 0.94] | 0.85* [0.78, 0.89] | **0.92 [0.88, 0.95]** | 6.0 [2.1, 8.3]×10⁻² | 13 [9.9, 16]×10⁻² | 0.46 [0.39, 0.52] | 0.48 [0.44, 0.51] |

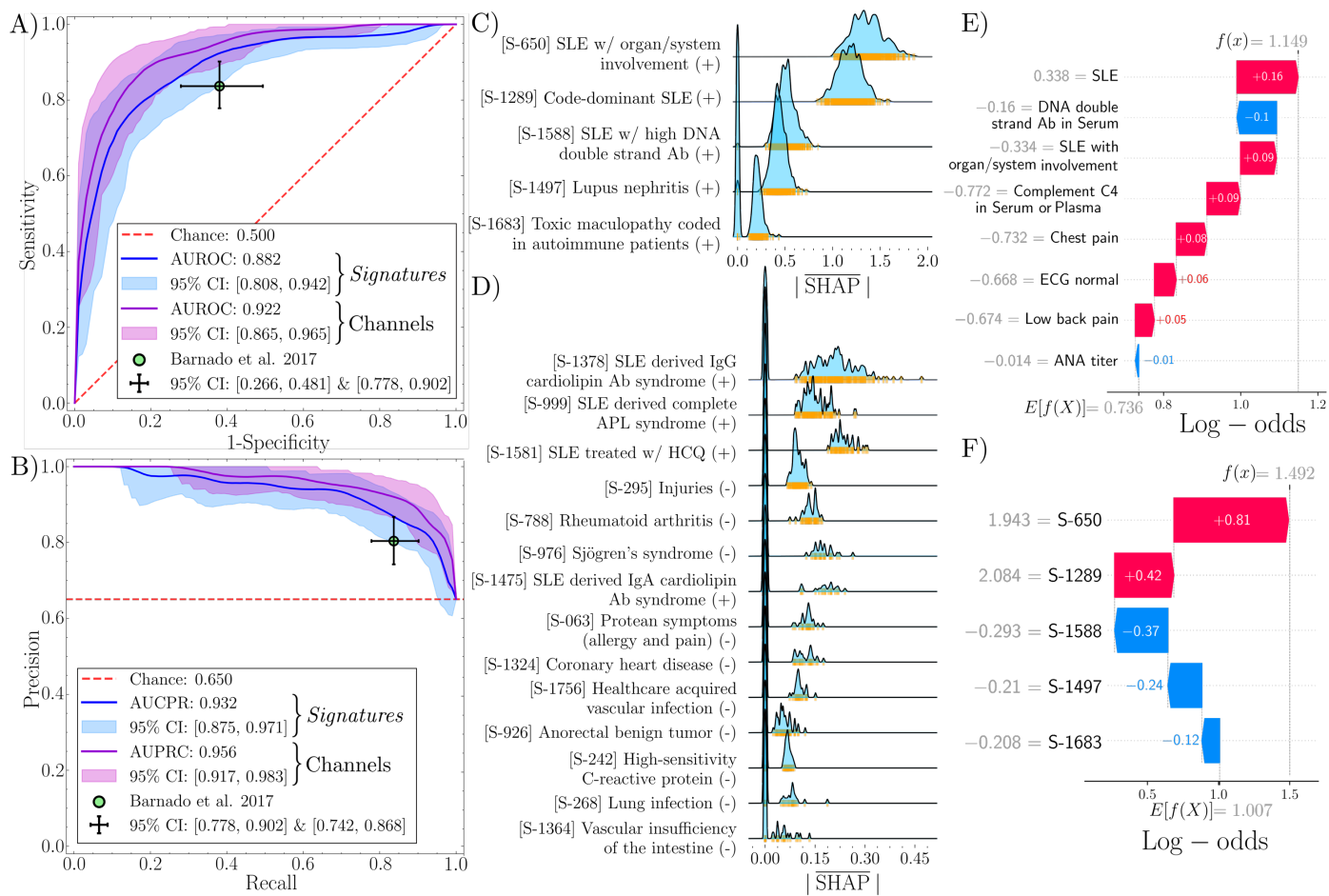

**Figure 4.** Test set **A)** receiver operating characteristic curve and **B)** precision-recall curve for both final models. Green dot shows the performance of Barnado et al.[17]. **C)** Mean absolute SHAP value (in log-odds units) empirical distribution of the nonzero features for the **C)** 5 most predictive signatures that make up the final signatures model and **D)** the additional 14 nonzero sources found to be predictive under training set sampling variation. Y-axis symmetric logarithmic scale threshold and KDE bin width were adjusted source-wise to enhance visualization. Each source model coefficient sign is provided in parenthesis to indicate the directionality of its effect in the prediction (−: away from SLE label). SHAP waterfall diagrams for the **E)** channels and **F)** signatures final models explanations for the same patient. Y-axis gray numbers indicate the feature value. Arrows show each predictor SHAP value which quantify the marginal contribution to model output ($f(x)$) starting from the expected estimate over the training set ($E[f(X)]$). APL: Antiphospholipid syndrome; HCQ: Hydroxychloroquine.

Sample variability testing revealed 14 additional predictive sources (figure 4D) most of which have signatures that are recognizable clinical patterns in SLE care (diagrams not shown for space). We identified three distinct isotypes of antiphospholipid syndrome, complete[16] (S-999), and anticardiolipin IgG (S-1378) and IgA (S-1475)[51]. The model also recognized sources for the two main autoimmune conditions that overlap with SLE, RA (S-788) and Sjögren's syndrome (S-976), the expression of which moves the prediction away from SLE. Treatment of SLE with hydroxychloroquine (S-1581) was identified as predictive. Increased expression of the 'Injuries' source (S-295) moved the prediction away from SLE, which aligns with the lifestyle of these patients.

**Record-specific prediction interpretability.** Figure 4E and 4F display the SHAP explanations for the same "near miss" patient by the channels and signatures models. Clinical note review revealed a long history (>30 years) of autoimmune diagnoses with overlapping RA and SLE and clinician disagreement. Assessment by rheumatologist confirmed this was a case of cutaneous lupus with no evidence of systemic damage to qualify as SLE. It is an example of the type of situation where higher interpretability at patient-level may be more useful than accuracy. Even though both models give a wrongly confident prediction that this is an SLE record, the explanations differ in both the depth and clinical relevance of the information they offer. Small negative expressions of the SLE laboratory pattern (S-1588), lupus nephritis (S-1497), and long-term user of hydroxychloroquine (S-1683) signatures pull the prediction toward the correct label. However, the high intensity of SLE codes, captured by the expressions of S-650 and S-1289, misleads the model to assign this record $P(SLE) = 0.82$. Conversely, the channels' view does not provide as detailed of an explanation even though it uses more variables. It does capture the high intensity of SLE codes, but also conveys a mix of codes and laboratory values that may be misleading. Note that almost all of them, lead the model to wrongly assign a $P(SLE) = 0.76$ to this patient.

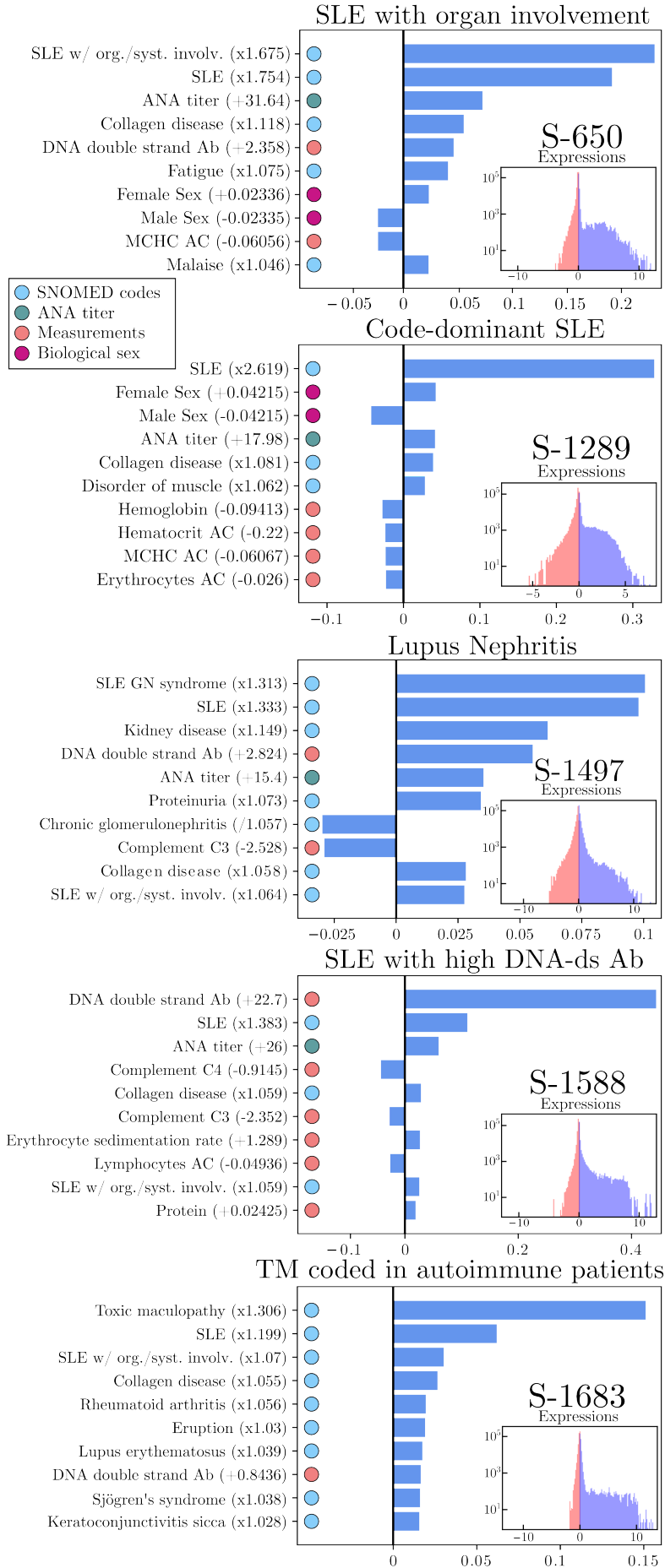

**Figure 5**. Signatures description diagrams for the top 5 most predictive sources. Bar length indicates the normalized change of a channel per unit of expression. This change is shown inside parenthesis in the original data space. For example, if a patient expresses 10 of S-650 the record would experience a 75.4% change on SLE code intensity which corresponds to a factor of $1.754^{10} = 275.6$, an increase in ANA titer of $10 \times 31.64 = 316.4$ and $10 \times 0.023 = 0.23$ higher probability of female sex. Diagrams show the top 10 contributor channels ordered by normalized change. Insets are log-scaled histograms of expression levels for all cross-sections in the discovery set. Expression units are individually scaled for each signature such that the standard deviation is 0.5, placing 95% of all expressions within the interval [-1, 1].
SLE w/ org./syst. involv.: SLE with organ/system involvement; GN: Glomerulonephritis; DNA-ds Ab: DNA double stranded antibody test; TM: Toxic maculopathy.

**Discussion and Conclusions**

We proposed a data-driven approach to address the problem of recognizing SLE's heterogeneity in the health record. We discovered 2000 independent latent sources and its corresponding EHR signatures from large amounts of unlabeled, noisy, sparse, and high-dimensional data from rheumatology patients. ICA-inferred sources represent the most-upstream causes of the SLE label within the data generating process insofar as the observed data permits their identification. Assuming a linear structural equation model the sources are the predictive exogenous error terms of each patient causal graph. Using the *source expressions* as input data representation forces supervised models to exploit the causal paths that generated the SLE labels which might be more predictive, interpretable, and clinically informative

We tested the sources against the original 7947 clinical variables (or channels) by training analogous SLE recognition models. Under internal validation the latent sources provide supervised models with higher model interpretability, more clinically relevant explanations and calibration although slightly lower discrimination. Notably, both models outperformed an optimistic view of the current go-to SLE phenotyping algorithm.

One limitation of our study is the lack of a bijective relationship between the observed variables and the sources. Having less sources than channels (2000 <

7947) implies that some of the sources may be linear combinations of error terms in the causal graph. If this were the case, the sources predictivity will be obscured and that would explain our findings. However, this limitation does not invalidate the causal value and higher interpretability of the sources *vs.* the channels in recognizing SLE heterogeneity.

The sources were discovered with no human input. We *let the data tell us* what linear combinations of changes in clinical variables define each signature. These independent patterns seem to reflect what clinicians look for when making sense of a record. And these may be known or unknown and current or emergent. We identified 19 clinical signatures had high clinical face validity and defined meaningful factors of SLE heterogeneity in our training set. We suspect that the true heterogeneity of SLE is likely greater, with more than 19 latent sources. Our study was limited by our learning set which was mainly of white race and not representative of the SLE population at large. It is well-known that SLE has a higher prevalence and severity among minority populations, thus it is reasonable to expect that their inclusion would reveal additional latent sources. This is a promising line of future work.

At the patient level, providers may be willing to trade off a small amount of accuracy for more meaningful causal explanations that can quantify the why behind their specific patient prediction. The equivalence between the ICA sources and error terms in the patient-specific causal graph allows to use SHAP as a framework to both quantify the predictivity of each source and explain the signatures model predictions making these easier to comprehend compared to those from the channels model. The signatures model is also more interpretable which is especially useful for challenging cases where the clinical history is not very informative, incoherent or reflects several diagnoses of overlapping conditions. In such scenarios, a provider would benefit from causal explanations that can be used to audit the model and ensure that its behavior complies with domain knowledge before making an interventional decision.

**Acknowledgements**
MBM received fellowships from Fulbright Spain and "la Caixa" Foundation (ID: 100010434, code: LCF/BQ/EU22/11930087). This project was funded by NIAMS/NIH R01 AR076516, Lupus Research Alliance BMS Accelerator Award, and CTSA award UL1 TR002243 from the National Center for Advancing Translational Sciences.